\pdfoutput=1

\documentclass[11pt]{article}

\usepackage[]{EMNLP2023}

\usepackage{times}
\usepackage{latexsym}

\usepackage[T1]{fontenc}

\usepackage[utf8]{inputenc}

\usepackage{microtype}

\usepackage{inconsolata}

\usepackage{graphicx,xcolor}
\usepackage{url}

\graphicspath{ {./pictures/} }

\title{Teach Me How to Improve My Argumentation Skills: \\ A Survey on Feedback in Argumentation}

\author{
   Camélia Guerraoui${}^{1,2,3}$\quad
   Paul Reisert${}^{4}$\quad
   Naoya Inoue${}^{5,2}$\quad
   Farjana Sultana Mim${}^{7}$\\
   {\bf Shoichi Naito${}^{1,2,6}$\quad
    Jungmin Choi${}^{1,2}$\quad
    Irfan Robbani${}^{5}$\quad
   Wenzhi Wang${}^{1,2}$\quad
   Kentaro Inui${}^{1,2}$}\\
    ${}^{1}$Tohoku University\quad
    ${}^{2}$RIKEN\quad
    ${}^{2}$INSA Lyon\quad
    ${}^{4}$Beyond Reason\quad \\
    ${}^{5}$JAIST\quad
    ${}^{6}$Ricoh Company, Ltd.\quad
    ${}^{7}$Tufts University\\ 
    {\small \texttt{\{guerraoui.camelia.kenza.q4, naito.shoichi.t1, wang.wenzhi.r7\}@dc.tohoku.ac.jp}, beyond.reason.sp@gmail.com} \\
    {\small \texttt{naoya-i@jaist.ac.jp, farjana.mim@tufts.edu, jungmin.choi@riken.jp, robbaniirfan@jaist.ac.jp, kentaro.inui@tohoku.ac.jp}}
}

\begin{document}
\maketitle
\begin{abstract}

The use of argumentation in education has been shown to improve critical thinking skills for end-users such as students, and computational models for argumentation have been developed to assist in this process. Although these models are useful for evaluating the quality of an argument, they oftentimes cannot explain why a particular argument is considered poor or not, which makes it difficult to provide constructive feedback to users to strengthen their critical thinking skills.
In this survey, we aim to explore the different dimensions of feedback (Richness, Visualization, Interactivity, and Personalization) provided by the current computational models for argumentation, and the possibility of enhancing the power of explanations of such models, ultimately helping learners improve their critical thinking skills.

\end{abstract}

\section{Introduction}

Argumentation is the field of elaboration and presentation of arguments to debate, persuade, and agree, where an argument is made of a conclusion (i.e., a claim) supported by reasons (i.e., premises)~\cite{walton2008argumentation}. By analogy with computational linguistics, \textit{computational argumentation} refers to the use of computer-based methods to analyze and create arguments and debates~\cite{gurevych-etal-2016}. It is a subfield of artificial intelligence that deals with the automated representation, evaluation, and generation of arguments. This field includes important applications such as mining arguments~\cite{al-khatib-etal-2016-cross}, assessing an argument's quality~\cite{el-baff-etal-2018-challenge}, reconstructing implicit assumptions in arguments~\cite{habernal-etal-2018-argument} or even providing constructive feedback for improving arguments~\cite{naito-etal-2022-typic}, to name a few.

In the context of education, learning argumentation (e.g., writing argumentative essays, debates, etc.) has been shown to improve students' critical thinking skills~\cite{pitchers-sodden-2000, behar-horenstein-etal-2011-teaching}. To further improve critical thinking skills, several researchers have been working on computational argumentation to support and provide tools to assist learners in improving the quality of their arguments.

Although computational models for argumentation are proven to assist students' learning and reduce teachers' workload~\cite{twardy-2004, wambsganss-etal-2021-arguetutor}, such models still lack the ability to \emph{explain} how an argument can be improved efficiently; e.g., why a particular argument was labeled bad or given a low score by their automatic evaluation rubrics. In other words, the model should be not only able to provide its results but also be able to \textit{explain and visualize the results in a comprehensive way} for the users so that users can understand, and ultimately improve their argumentation skills.

\begin{figure}[t]
    \centering
    \includegraphics[width=\linewidth]{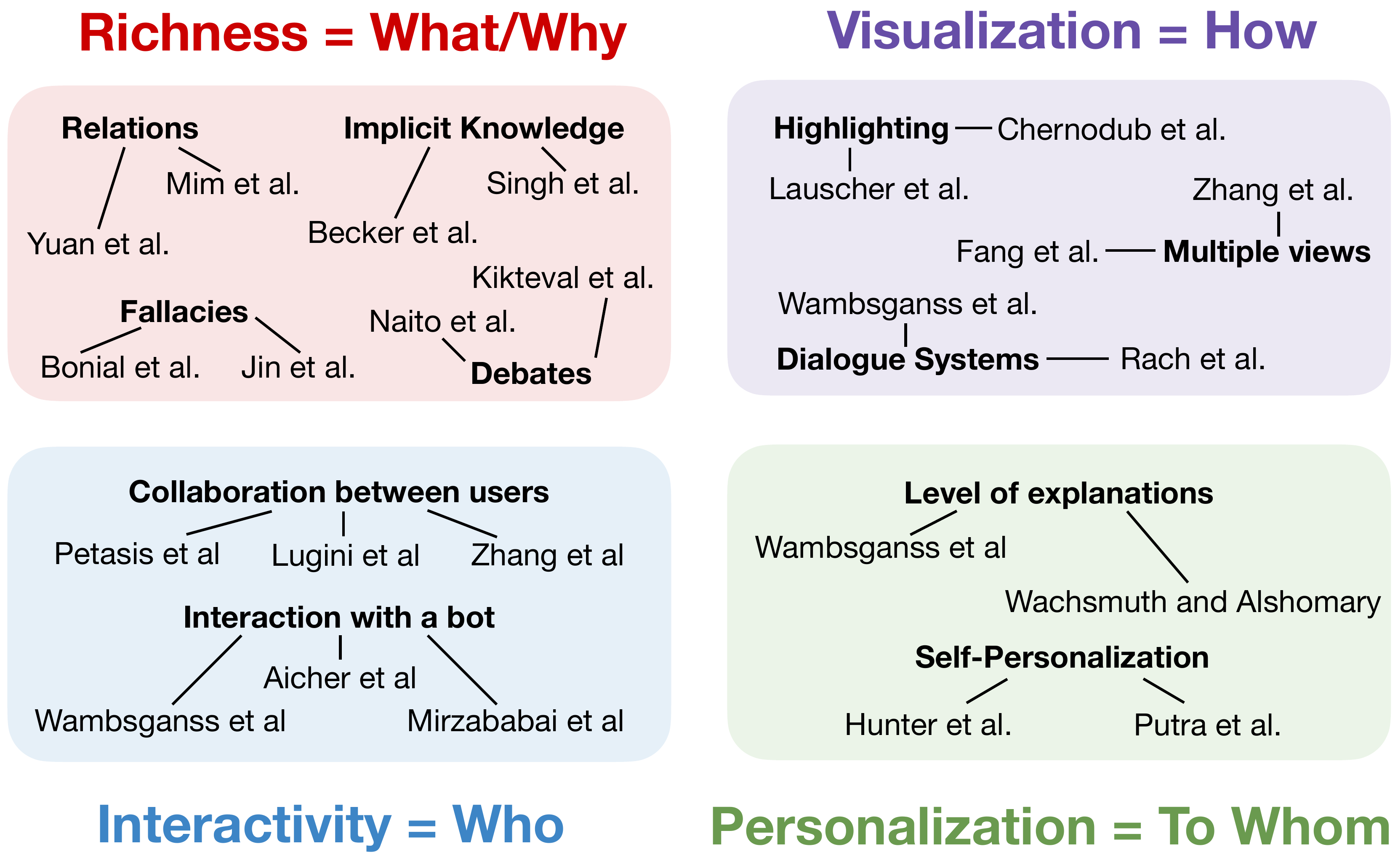}
    \caption{Overview of works focusing on the evaluation and improvement in argumentation.}
    \label{fig:overview}
\end{figure}

We argue that the output for current computational models for argumentation act as a type of explanation and must be the focus of future work. For our survey, we categorize works into four different dimensions (cf., Figure~\ref{fig:overview}):
\begin{itemize}
    \item \textit{Richness}: Level of feedback details given by a model, i.e., \textit{what} is the error identified by the model and \textit{why} it is an error;
    \item \textit{Visualization}: Way of presenting feedback, i.e., \textit{how} the explanation is shown;
    \item \textit{Interactivity}: Ability to communicate with the model, other users, or a third-person, i.e., with \textit{whom} the user is talking;
    \item \textit{Personalization}: Ability to adapt the feedback to the users' background, i.e., \textit{to whom} the feedback is given.
\end{itemize}

\begin{figure}[t]
    \centering
    \includegraphics[width=\linewidth]{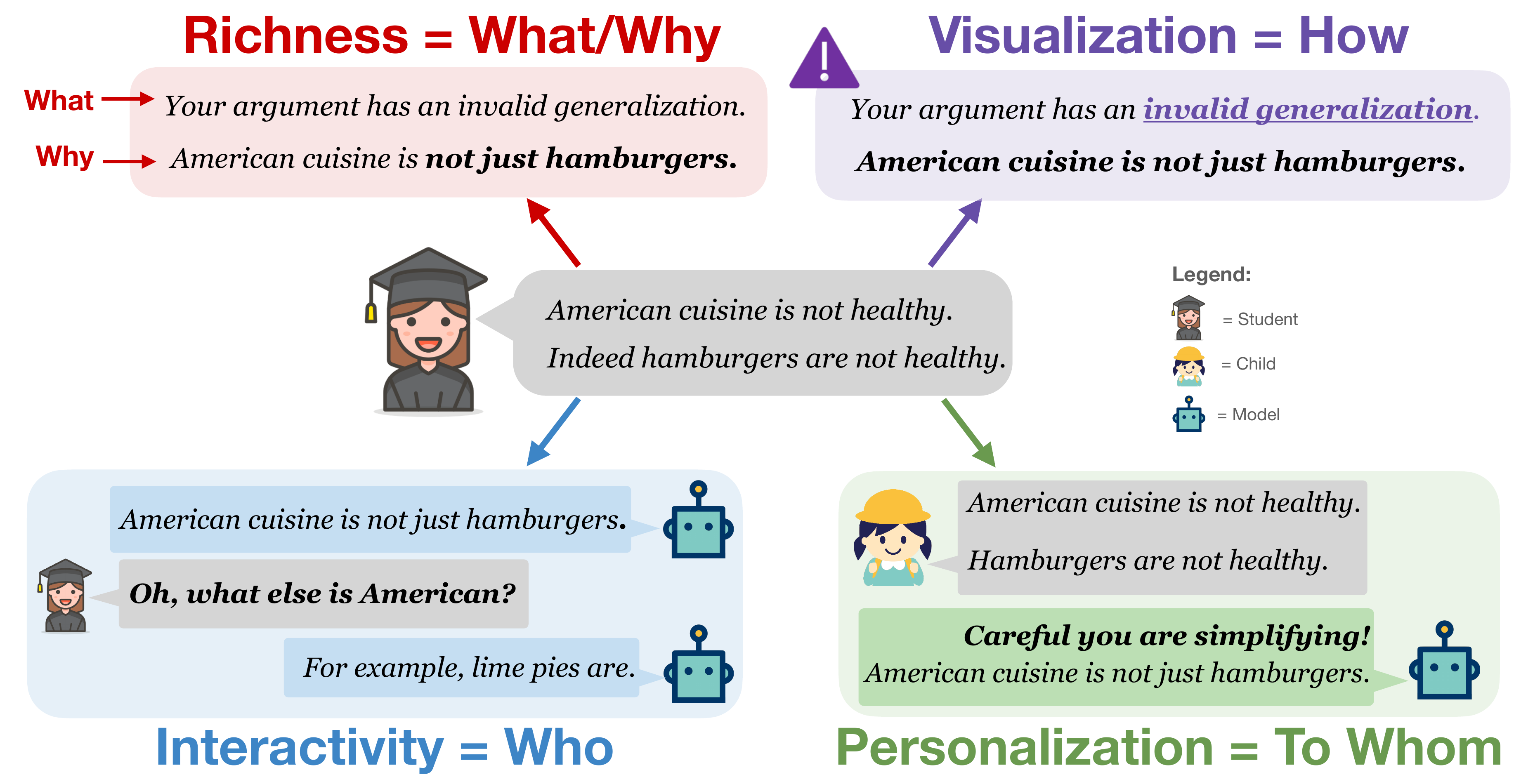}
    \caption{Example of explanations in the dimensions \textit{Richness}, \textit{Visualization}, \textit{Interactivity} and \textit{Personalization} for the same input argument.}
    \label{fig:ex-types}
\end{figure}

In Figure~\ref{fig:ex-types}, for a given argument consisting of two claims and one premise, four different feedback are shown, each highlighting a different dimension of feedback (\textit{Richness}, \textit{Visualization}, \textit{Interactivity}, and \textit{Personalization}).

Towards explainable computational argumentation, this survey aims to give an overview of computational argumentation on automated quality assessment.
We explore work providing explanations answering the following: \textit{What} (\S\ref{sec:richness-what}), \textit{Why} (\S\ref{sec:richness-why}), \textit{How} (\S\ref{sec:visualization}), \textit{Who} (\S\ref{sec:interactivity}), and \textit{To Whom} (\S\ref{sec:personalization}).
Finally, we discuss remaining challenges and potential ways to overcome them (\S\ref{sec:open_issues}) in order to develop systems that provide explanations in a way in which learners can improve their critical thinking skills.
We believe this survey can aid researchers in understanding current explanations in argumentation and broaden their horizon on argumentative feedback.\footnote{\label{foot:website}For more details, papers mentioned in this survey are categorized at \url{https://anonymized}.}


\section{Related Work}





Several surveys have been done in the field of argumentation \cite{ke_2019,beigman-klebanov-madnani-2020-automated,lawrence_chris_2020,xinyu_2022} and explainability \cite{danilevsky_2020,islam_2021,hartmann_2022}. In this section, we focus on the recent surveys related to explainability in argumentation.

First, ~\citet{vassiliades_2021} highlight the potential of argumentation in explainable AI systems. They provide an exhaustive overview of argumentation systems by grouping them by domain, such as law, medicine, and semantic web. For each domain, papers are compared by tasks (e.g, argument classification). decision-making  justification, explanation through dialogue, and argument classification. Despite the extensiveness of this survey, some topics important for improving explanations in argumentative systems received little attention. For example, frameworks that include arguments with commonsense knowledge and diverse attack relations between them
have rarely been discussed, even though, they can enhance the model's explainability~\cite{saha-etal-2021-explagraphs}.

~\citet{cyras_2021} focus on the different frameworks, types, and forms of explanations. They distinguish intrinsic approaches (models using argumentative methods) from post-hoc approaches (non-argumentative models that provide complete or partial explanations). They discuss multiple forms of argumentation, such as dialogue, extensions, and sub-graphs. Their final roadmap is an essential part of their work and covers the need to focus more on properties and computational aspects of argumentation-based explanations. Whereas they focus on argumentation used to explain, our work discusses how computational argumentation needs more explanation.

Moreover, our work distinguishes itself from the two previous surveys~\cite{vassiliades_2021, cyras_2021} by focusing on evaluating and improving users' critical thinking skills.
\section{Pedagogy}
\label{sec:pedagogy}

Based on ~\citet{rapanta-etal-2013-competence}, researchers do not agree on a uniform definition of argumentation competence, nor is there a universally accepted method or guideline to analyze and evaluate the various components of argumentation competence, making it challenging for NLP researchers to choose a pedagogical method to use to generate feedback. This section presents some standard pedagogical methods used in teaching argumentation.

\paragraph{Toulmin model}
The Toulmin model, often seen as the founding of teaching argumentation, is a popular framework for analyzing, constructing, and evaluating arguments, can contribute to the improvement of students’ argumentative writings ~\cite{rex-etal-2010-toulmin, yeh-1998-toulmin} as well as critical thinking skills~\cite{giri-2020-toulmin}. This approach deconstructs an argument into six elements: claim, data, warrant, backing, qualifier, and rebuttal, and students are taught to identify each element within an argument. 

\paragraph{Collaborative argumentation}
In collaborative argumentation-based learning, also described as CABLE by \citet{baker-etal-2019-collabo}, individuals or groups work together to construct, refine, and evaluate arguments on a particular topic or issue. The main goal of collaborative argumentation is to foster constructive dialogue, critical thinking, and the exploration of different perspectives.

~\citet{weinberger-fischen-2006-collabo} differentiate four dimensions of CABLE:
\begin{itemize}
    \item \textit{Participation}: Do learners participate at all? Do they participate on an equal basis? 
    \item \textit{Epistemic}: Are learners engaging in activities to solve the task (on-task discourse) or rather concerned with off-task aspect?
    \item \textit{Argumentative}: Are learners following the structural composition of arguments and their sequences?
    \item \textit{Social}: To what extent do learners refer to the contributions of their learning partners? Are they gaining knowledge by asking questions?
\end{itemize}

~\citet{veerman-etal-2002-collabo, baker-etal-2019-collabo} show the positive effects of CABLE on students' development of argumentation. 

\paragraph{Tree-based feedback}
A customary approach to giving feedback to students is through the provision of free text feedback. Text-based feedback allows for more in-depth analysis and can be particularly useful when providing specific examples or referencing external resources.
Nevertheless, it is worth noting that generating free-text feedback can be time-consuming. Conversely, the advent of tools like ~\cite{putra-etal-2020-tiara} has given rise to tree-based feedback. 
This feedback form employs graphical representations, such as concept maps or mind maps, to illustrate the relationships between ideas or concepts. This visual approach can help students visualize the connections between different concepts and enhance their understanding of complex topics ~\cite{matsumura-sakamoto-2021-tree}.

\paragraph{Socratic questioning}

The Socratic questioning is a teaching strategy commonly used in education, where the student is guided, through reflexive questions, towards solving a problem on their own, instead of being given the solution directly described in ~\cite{schauer-2012-socratic, abrams-2015-socratic}.
Recently, this method has been integrated into Large Language Models (LLMs) to more effectively adhere to user-provided queries ~\cite{ang-etal-2023-socratic, pagnoni-etal-2023-socratic}, to enhance the ability of such models in generating sequential questions~\cite{shridhar-etal-2022-automatic}, but also enhancing the explainability of these models ~\cite{al-hossami-etal-2023-socratic}.

Nevertheless, the Socratic questioning is now raising debates among researchers focusing on pedagogy in argumentation. Indeed ~\citet{kerr-1999-socratic, christie-2010-socratic} pointed out its inefficiency and abusiveness as students are forced to give imperfect answers in a hurry and endure criticism.
\section{Richness - What is an Error?}
\label{sec:richness-what}

To improve students' critical thinking skills, we first need to evaluate their argumentative texts, i.e., identify argumentative errors. In this section, we focus on models providing shallow explanations, i.e., models that identify \textit{what} should be corrected in the arguments. We discuss recent works that identify properties such as the structure of arguments helpful to assist in this process.

\paragraph{Components}
Identifying argumentative components is one of the fundamental tasks in argumentation~\cite{teuful-zoning-1999,stab-gurevych-2014-identifying, emnlp-2020-jo}. Such works primarily focus on identifying components such as \textit{claims} and \textit{premises}.
More recently, the usefulness of identifying such components can be seen in tasks such as counter-argument generation. For example, in ~\citet{alshomary-etal-2021-counter}, weak premises are identified and ranked in order to generate counter-arguments.

\paragraph{Relations}
After identifying the different components of an argumentative text, it is necessary to distinguish the multiple relations between them to assert the quality of the arguments' quality. Indeed, supporting or refuting a claim is made of complex logical moves, such as promoting, contradicting, or acknowledging a fact. Therefore it is not trivial to use correct logic. To identify the different relations patterns, ~\citet{yuan-etal-2021-leveraging} focus on finding interactive argument pairs, whereas ~\citet{mim-etal-2022-lpattack} enables annotating complex attack relations.

\paragraph{Schemes}
In addition to components and relations, ~\citet{walton2008argumentation} proposed a set of roughly 80 logical argumentation schemes to categorize the underlying logic. Each scheme has a set of critical questions which provide a template to assess the strength of the argument depending upon the associated scheme. Since the first work on automatically detecting argumentation schemes in argumentative texts~\cite{feng-hirst-2011-classifying}, the use of such schemes has been explored in tasks such as essay scoring~\cite{song-etal-2014-applying}.

\paragraph{Fallacies}
Although a good structure with a claim and premises is necessary for a good argument, it is not sufficient. Indeed an argument has other more complex properties, such as its logical, dialectical, and rhetorical aspects.
A fallacy is a logical error or deceptive argument that undermines the validity of a conclusion or reasoning, which poses a substantial issue due to its propensity to generate miscommunication.
Towards teaching students to avoid making errors in logical reasoning, logical fallacies have received attention~\cite{habernal-etal-2017-argotario,bonial-etal-2022-search, sourati-etal-2023-fallacy}.
Motivated by the gamification method made by Habernal et al., Bonial et al. aimed to capture similar fallacy types for news articles, but the low distribution of fallacy types in the wild makes identification challenging.
Indeed most natural texts do not have recurrent specific patterns, compared to current datasets, like the Logic and LogicClimate datasets~\cite{jin-etal-2022-logical}.
Moreover, given the large number of logical fallacies that exist (over 100 types), long arguments can group multiple fallacies, resulting in difficulties in classification \cite{goffredo-etal-2022-fallacy}.

\paragraph{Debates}
In a case of a debate, an opponent is willing to give a counter-argument synchronously and interactively. Analyzing and evaluating a debate is a difficult task as we need to retrieve not only the argumentation structure of each opponent but also the relations between them.
~\citet{bao-etal-2022-arguments} focuses on argument pair extraction (APE), which consists of finding two interactive arguments from two argumentative passages of a discussion. Although the APE task gives insights into relations between different argumentative texts, it does not indicate complex relations (i.e., how claims, supports, attacks and the intention of the speakers are interrelated). To palliate this issue, ~\citet{hautli-janisz-etal-2022-qt30} identified and analyzed the dialogical argumentative structure of debates using Inference Anchoring Theory (IAT)~\cite{budsziyska2014model}.
Following the same IAT theory, ~\citet{kikteva-etal-2022-keystone} investigated the role of different types of questions (e.g., pure, assertive, and rhetorical questions) in dialogical argumentative setting and showed that different type of question leads to different argumentative discourse. Focused more on the opponent's side of a debate, ~\citet{naito-etal-2022-typic} propose diagnostic comments for assessing the quality of counter-arguments by providing expressive, informative and unique templates. The comments are then written by template selection and slot filling.

Although the identification of such argumentative structures (components, relations, and schemes) and properties (fallacies and debates pattern) is important, it has limitations in terms of effective feedback. Identifying a missing claim or a wrong premise is not enough to properly understand how to improve the argumentation. Therefore we relate the identification of structure and properties to shallow explanations in the sense that end-users can still benefit from the output of the models.




\section{Richness - Why is This an Error?}
\label{sec:richness-why}

Although shallow explanations help end-users to identify their mistakes, they tend to be minimalist and need more guidance. Shallow explanations can be hard to understand, specially for beginners in argumentation. To explain more effectively the errors in an argument, a model should go a step further, hence by providing \textit{in-depth} explanations, which attempt to identify the argument's implicit components to explain \textit{why} there is an error in an argument.

\paragraph{Implicit Knowledge and Reasoning in Arguments}
To provide \textit{in-depth} explanations, we need to know how to refine the argument, i.e., how to identify implicit information. Recently many works have focused their attention on this aim.
The main goal of such studies is to make the structure and reasoning of arguments explicit to better explain the arguments for humans. 
Additionally, this focus can eventually help build Robust Argumentation Machines that can be enriched with language understanding capacity. 
The ExpLAIN project .~\citet{becker-etal-2021-reconstructing} and ~\citet{jo-2021-kenli} are one such example that focuses extensively on reconstructing implicit knowledge in arguments by relying on knowledge graphs among others. 
Taking a step further in this direction, ~\citet{singh-etal-2022-irac} proposed to utilize such implicit information to bridge the implicit reasoning gap in arguments to help students explain their arguments better.


\section{Visualization - How to Show the Error?}
\label{sec:visualization}

The effectiveness of any argument does not solely rely on its content but also on its presentation.
This is where the art of visualization of argumentative feedback emerges as a crucial factor.
Visualizing feedback empowers individuals to perceive the intricacies of an argument in a more comprehensive and accessible manner.
By using visual aids like graphs and charts, feedback becomes more accessible and engaging, fostering constructive discussions.
In this section, we will see how visualization impacts argumentative feedback.

\paragraph{Highlights}

A first simple approach of visualization is highlighting, i.e., application of visual emphasis on specific pattern with the intention of drawing the viewer's attention on this specific pattern. For example, ~\citet{lauscher-etal-2018-arguminsci} identify the argument component (Claim, background, data) and visualizes them by highlighting the text in different colors. Similarly, ~\citet{chernodub-etal-2019-targer} allow the user to choose the model to use and the components to highlight. ~\citet{wambsganss-etal-2022-highlight} take a step further by presenting scores giving a quick overview of the user's skills. 

Highlighting serves as an essential key step in the cognitive input process, enabling viewers to quickly identify crucial argumentative structure. However, its use should be complemented with other visualization techniques to ensure a more profound exploration and comprehension of complex explanations. Studies conducted by ~\citet{lauscher-etal-2018-arguminsci, chernodub-etal-2019-targer, wambsganss-etal-2022-highlight} shed light on the potentials and limitations of highlighting, paving the way for future advancements in data visualization methodologies.

\paragraph{Multiple views}

To overcome the shallowness of highlighting, several researchers add not only a text editor to their system but also other views such as diagrams showing the argumentative structure. 
For example to compare two drafts of an essay, ~\citet{zhang-etal-2016-argrewrite, afrin-etal-2021-visualization} use a revision map made of color-coded tiles, whereas ~\citet{putra-etal-2021-tiara2} rely on a tree to reorder arguments.

Based on the work of ~\citet{wambsganss-etal-2020-al}, ~\citet{xia-etal-2022-persua, wambsganss-etal-2022-alen} use a text editor which highlights components, a graph view which shows an overview of the argumentative structure, and a score view showing the user's performance. Based on the classroom-setting evaluation, students using such systems wrote texts with a better formal quality of argumentation compared to the ones using the traditional approach. Nevertheless, the current accuracy of such systems' feedback still leave a large improvement space in order for users to be motivated to use them.

More recent work such as ~\citet{zhang-etal-2023-visar} incorporate feedback generated by state-of-the-art LLMs in their graphical systems. Nonetheless, factual inaccuracies, inconsistent or contradictory statements are still generated, exposing the user to confusion and leaving room for improvement.

\paragraph{Dialogue Systems}

In the realm of visualization, a novel approach gaining traction is the integration of dialogue systems to enhance the interaction between users and visual representations. Dialogue systems, commonly known as chatbots like ChatGPT, have been increasingly explored for their potential to improve critical thinking skills and facilitate information comprehension~\cite{rach-etal-2020-evaluation, wambsganss-etal-2021-arguetutor}. 

This kind of representation is challenging in term of the application's user-friendliness. Indeed, in an pedagogical context, it can be hard to track and visualize the user's progress. The user may also have difficulties in finding his previous assignments.

Despite the growing popularity of both graphs and chatbots in data visualization, limited work has directly compared their effectiveness in improving critical thinking skills. Further research is needed to provide more nuanced insights on the comparison on one hand between both approaches, on the other between works among the same approach. 

The importance of visualization in argumentative feedback lies in its ability to enhance the presentation and understanding of complex ideas.
This introductory section delved into the significance of visualization in argumentative feedback, highlighting its potential to improve students' learning process.
\section{Interactivity - Who Talks to the User?}
\label{sec:interactivity}

Teaching argumentation is a multifaceted task that demands more than the dissemination of theoretical knowledge; it requires fostering interactive learning environments that facilitate active engagement and practice. The traditional approach to teaching argumentation often centers on lecturing and one-way communication, where instructors impart information to students. While didactic methods have their place in education, a more interactive pedagogical approach, one that encourages learners to actively participate, can be used. In this section we will see in which ways current argumentative computational models enable a form of interaction. 

\paragraph{Interaction between different users}

NLP systems mostly allow communication between a user and a conversational agent. Nonetheless, some works chose to apply the CABLE pedagogical method (cf section (\S\ref{sec:open_issues})) and allow a user to dialog with other users. Following the footsteps of \citet{petasis-2014-nomad}, \citet{lugini-etal-2020-discussion} track real-time class discussions and help teachers annotate and analyze the discussions.

Even if recent research works such as ~\citet{zhang-etal-2023-visar} plan to add a collaborative setting, we realize through our survey that only a few works focus on collaboration between multiple users. 
The concept of collaboration between multiple users within NLP systems is promising. However, it is essential to acknowledge that some challenges and barriers have hindered its widespread adoption in research works, possibly due to the difficulty of designing and evaluating such tools, as they require important human resources. 

\paragraph{Interaction with a conversational agent}

As seen in the section (\S\ref{sec:visualization}), several research papers have showcased the feasibility of employing current conversational agents for educational purposes ~\cite{lee-etal-2022-bot, macina-etal-2023-bot, wang-etal-2023-bot}. Often based on state-of-the-art language models, these agents have shown great capabilities in understanding and generating human-like responses. They can engage in dynamic and contextually relevant conversations, making them potentially valuable tools for educational purposes.

The use of conversational agents as dialog tutors has been explored outside of argumentation ~\cite{wambsganss-etal-2021-arguetutor, mirzababaei-pammer-2022-bot, aicher-etal-2022-towards}. For instance, in ~\citet{mirzababaei-pammer-2022-bot}, an agent examines arguments to determine a claim, a warrant, and evidence, identifies any missing elements, and then assists in completing the argument accordingly. ~\cite{wambsganss-etal-2021-arguetutor} create an interactive educational system that uses interactive dialogues to teach students about the argumentative structure of a text. The system provides not only feedback on the user's texts but also learning session with different exercises.

Research on chatbots in education is at a preliminary stage due to the limited number of studies exploring the application of effective learning strategies using chatbots. This indicates a significant opportunity for further research to facilitate innovative teaching methods using conversational agents~\cite{hwang-chang-2021-chatbotChallenges}. Indeed, pre-work, extraction, and classification of useful data remain challenging as the data collected are noisy, and much effort still has to be made to make it trainable \cite{lin-etal-2023-reviewChatbot}.

Future researchers must also account for ethical considerations associated with chatbots, including value-sensitive design, biased representations, and data privacy safeguards, to ensure that these interactive tools positively impact users while upholding ethical standards \cite{kooli-2023-chatbot}. 

Overall, integrating interaction in teaching argumentation is not merely a pedagogical choice but an essential requirement to cultivate adept arguers who can navigate the intricacies of argumentation. Therefore, we encourage researchers to consider this dimension in their future pedagogical systems.
\section{Personalization - To Whom is it For?}
\label{sec:personalization}

Even if the explanations mentioned in the sections(\S\ref{sec:richness-what}) and (\S\ref{sec:richness-why}) are a step towards good guidance, they are static, which can be problematic depending on the end-user. Indeed beginners or professionals in argumentation do not need the same amount of feedback. A child and an adult have different levels of understanding and knowledge. Therefore it is essential that a model knows \textit{to whom} it should explain the errors and hence adapts its output by providing \textit{personalized} explanations.

\paragraph{Levels of explanations}

A first approach to personalization is to discretize the different users' backgrounds into a small number of categories based on their level of proficiency in argumentation. For example, with the systems described in ~\citet{wambsganss-etal-2020-al, wambsganss-etal-2022-alen}, users can choose their level (Novice, Advanced, Competent, Proficient, Expert).

Although ~\citet{wambsganss-etal-2020-al, wambsganss-etal-2022-alen} propose different granularity levels of explanations, their study is restrained to students from their university.
Having end-users from different backgrounds may imply the need for new levels of explanations. Indeed, ~\citet{wachsmuth-alshomary-2022-mama} show that the age of the explainee changes the way an explainer explains the topic at hand. Information such as the learner's age should be considered in future interactive argumentative feedback systems, where terminology such as \textit{fallacy} and their existence would require different approaches of explanation for younger students (i.e., elementary) compared to older students.

\paragraph{Self-personalization}
For more personalized feedback, some systems such as ~\citet{hunter-metal-2019-personalization, putra-etal-2020-tiara} rely on the user's input.
For example, they allow users to make their custom tags or to choose their preferences among a set of rubrics. Nevertheless, letting the user manually personalize the system can be overwhelming and time-consuming for users. 

\paragraph{Next directions}

~\citet{hunter-metal-2019-personalization} argue that the next direction for personalized argumentative feedback would be to develop argumentation chatbots for persuasion and infer the user's stance based on the discussion. Chatbots' personalization capabilities enabling them to tailor their responses to individual learners' needs and learning styles, potentially enhancing the effectiveness of the tutoring process\cite{lin-etal-2023-reviewChatbot}. 
Bridging the gap between personalized chatbots \cite{qian-etal-2021-pchatbot, ma-etal-2021-chatperperson}, personalized educational methods \cite{gonzalez-etal-2023-persoEdu, ismail-etal-2023-persoSurvey} and argumentation has remained unexplored.

Therefore we think researchers should focus furthermore in the future on providing more \textit{personalized} explanations (i.e., precisely adjusted by considering the background of the learner) to efficiently improve the critical thinking skills of an end-user.
 
\section{General Open Issues}
\label{sec:open_issues}

Teaching argumentation through the use of NLP systems holds significant promise for enhancing educational experiences among students. However, the further research in this area still present various open issues. In this section, we explore the overall difficulties in designing and evaluating computational models for argumentation and discuss some methods for mitigating them.

\paragraph{Evaluating different systems}

The evaluation of NLP systems often heavily relies on human assessment, which is very insightful. However, this reliance on human evaluation makes it hard to reproduce the evaluation and compare different systems. To the best of our knowledge, no research have been dedicated to comprehensive comparative studies of different end-to-end systems.

While some systems, such as \citet{wambsganss-etal-2021-arguetutor}, exhibit promising performance, the lack of direct comparisons with other similar systems hampers the comprehension of their relative advantages and limitations. As researchers and educators, it becomes overwhelming to discern which system best fits specific pedagogical objectives.

One possible reason for this issue resides in the restricted access of various tools. Indeed, many systems may not be readily accessible as open-source resources, limiting researchers to test them.
Additionally, the lack of common guidelines with standardized metrics for evaluating NLP systems designed for teaching argumentation exacerbates the difficulty in evaluating different systems in a systematic manner. Metrics such as coherence, are essential aspects of an argumentation system's performance, but the absence of a uniform evaluation framework makes it harder to quantitatively compare different tools.

Therefore we encourage researchers to find a common guideline to evaluate their systems and to give access to their research.

\begin{figure*}[ht]
    \centering
    \includegraphics[width=0.9\textwidth]{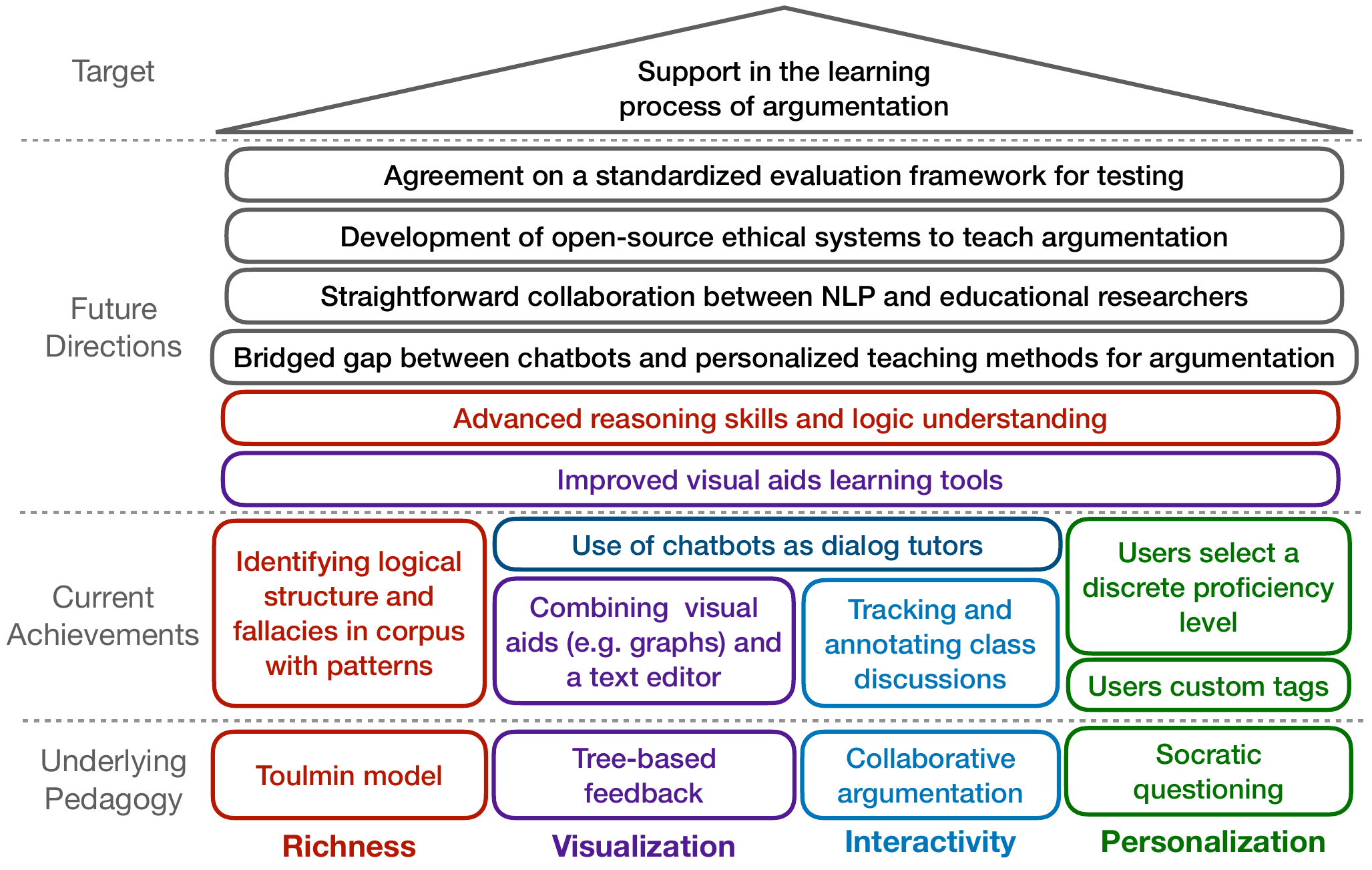}
    \caption{Current and future directions of teaching argumentation with NLP systems. Future directions with a specific color correspond to a specific dimension whereas the ones in black are general directions.}
    \label{fig:future}
\end{figure*}

\paragraph{Domain Adaptation}
Towards effectively explaining output in a way which can improve critical thinking skills of end-users, future systems must be capable of understanding the topic of discussion in such a way that argumentation errors (e.g., fallacies, etc.) can be identified. In a pedagogical setting, teachers have the ability to choose new topics of discussion annually; hence, systems must also be capable of adapting to various domains. Recent works have focused on the task of domain adaptation for tasks such as short answer scoring~\cite{funayama-etal-2023-aied}, which focus on training models for several tasks in order to learn common properties useful for evaluating unseen topics. For computational argumentation, we must also adopt such strategies to ensure the most reliable feedback is given to the end-user. 

\vspace{-5pt}
\paragraph{Need to simplify the collaboration process}
As mentioned a priori, NLP researchers and pedagogical researchers generally conduct their research independently, thus creating a gap. We argue that researchers from both fields must come together to ensure appropriate, sufficient explanations are provided to learners. However, presently it is difficult to find the appropriate learners for evaluation. Ideally, a system for linking various educational schools and providers with artificial intelligence researchers could significantly help assist with ensuring systems can be properly evaluated.

\paragraph{Ethics}
Ideally, personalizing a constructive feedback system for each user, specific to their background and current understanding of the world, would benefit the user significantly. However, due to ethical issues and complex implementation, focusing on certain groups for personalizing feedback should be focused on. For example, in the context of education, the way of explanation for models can be categorized by age group (i.e., elementary vs. high school students). When integrating LLMs into the classroom, we must be careful to confine to appropriate standards.
\section{Conclusion}

Our survey delves into the domain of argumentation in education and the utilization of computational models to enhance critical thinking skills among students. 
In this survey, we explored several works providing explanations in argumentation, following various dimensions of feedback: \textit{Richness} (\S\ref{sec:richness-what}, \S\ref{sec:richness-why}), \textit{Visualization} (\S\ref{sec:visualization}), \textit{Interactivity} (\S\ref{sec:interactivity}), and \textit{Personalization} (\S\ref{sec:personalization}).

We identified different potential areas for improvement to enhance the overall quality of educational systems to teach argumentation. We summarized these challenges with the following points: (1) generate accurate, constructive feedback for a real-life input, (2) tailor the output based on the user's background, (3) evaluate and compare more deeply end-to-end systems, (4) collaborate with pedagogical teams and actual students, and finally (5) take into consideration ethical issues.

For instance, in challenge (2), the use of conversational agents becomes increasingly frequent. Nevertheless, such systems still leave room for improvement; in particular, only a few research focuses on using their ability to tailor the discussion based on the user to improve critical thinking skills.

We hope our survey contributes to expanding the argumentation community's horizons with a comprehensive understanding of current perspectives in NLP systems for teaching argumentation.

In our future work, we decided to focus further on real-life user-friendly end-to-end systems (challenges (1) and (3)). We plan to prototype a system to measure the effects of different feedback on critical thinking skills and determine how different interfaces can impact the learning process by evaluating this system in actual classrooms (cf., Figure~\ref{fig:interface} in the Appendix~\ref{sec:appendix}).

\clearpage
\section*{Acknowledgements}
This work was supported by JSPS KAKENHI Grant Number 22H00524.

\bibliographystyle{acl_natbib}
\bibliography{bibliography/anthology, bibliography/custom}

\clearpage
\appendix

\onecolumn
\section{Appendix}
\label{sec:appendix}

\begin{figure}[ht]
    \centering
    \includegraphics[width=\textwidth]{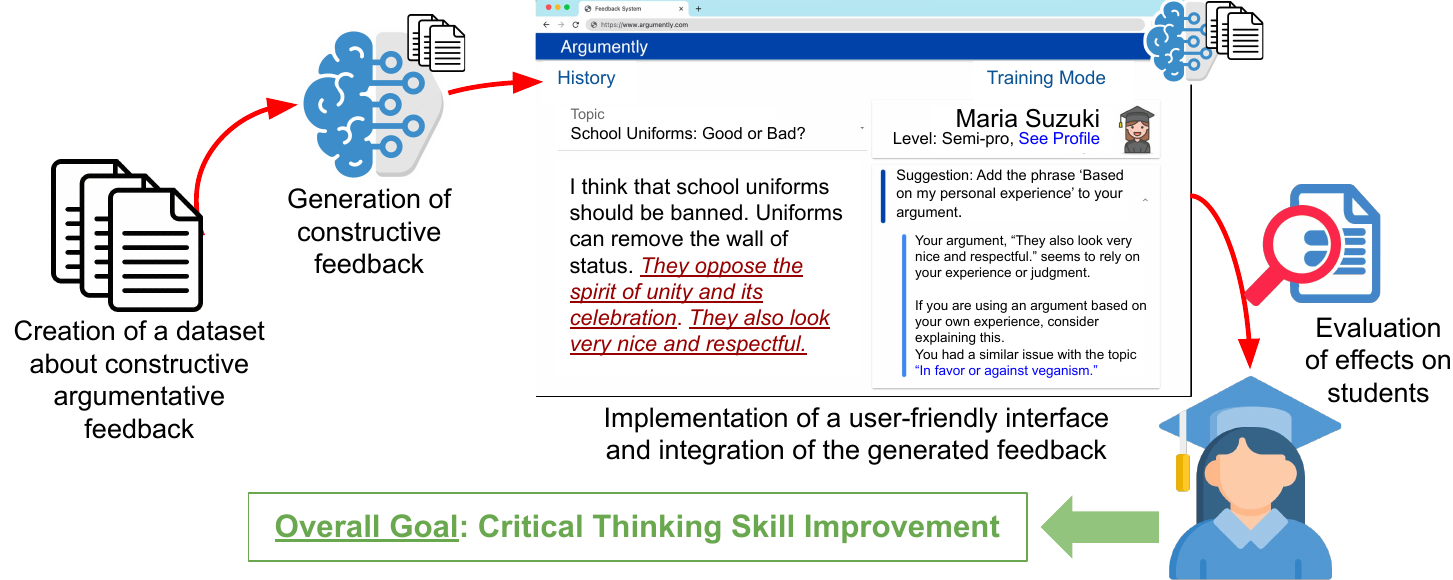}
    \caption{Preliminary sketch design of an end-to-end system to learn argumentation.}
    \label{fig:interface}
\end{figure}

\end{document}